\documentclass[10pt,twocolumn,letterpaper]{article}

\usepackage[accsupp]{axessibility}
\usepackage{iccv}
\usepackage{times}
\usepackage{epsfig}
\usepackage{graphicx}
\usepackage{amsmath}
\usepackage{amssymb}
\usepackage{enumitem}

\usepackage[ruled]{algorithm2e}
\usepackage{multirow}

\usepackage[pagebackref=true,breaklinks=true,letterpaper=true,colorlinks,bookmarks=false]{hyperref}

\iccvfinalcopy 

\def\iccvPaperID{1601} 

\ificcvfinal\fi

\begin{document}

\title{G2L: Semantically Aligned and Uniform Video Grounding \\via Geodesic and Game Theory}

\author{
Hongxiang Li$^{1}$, Meng Cao$^{2,1}$, Xuxin Cheng$^{1}$, Yaowei Li$^{1}$, Zhihong Zhu$^{1}$, Yuexian Zou$^{1}$\footnotemark[2]\\
$^{1}$School of Electronic and Computer Engineering, Peking University \\ $^{2}$International Digital Economy Academy (IDEA)\\
{\tt\small \{lihongxiang, chengxx, zhihongzhu, ywl\}@stu.pku.edu.cn; \{mengcao, zouyx\}@pku.edu.cn}
}

\maketitle

\renewcommand{\thefootnote}{\fnsymbol{footnote}}

\footnotetext[2]{~Corresponding author.}

\ificcvfinal\thispagestyle{empty}\fi

\begin{abstract}
    The recent video grounding works attempt to introduce vanilla contrastive learning into video grounding. However, we claim that this naive solution is suboptimal. Contrastive learning requires two key properties: (1) \emph{alignment} of features of similar samples, and (2) \emph{uniformity} of the induced distribution of the normalized features on the hypersphere. Due to two annoying issues in video grounding: (1) the co-existence of some visual entities in both ground truth and other moments, \ie semantic overlapping; (2) only a few moments in the video are annotated, \ie sparse annotation dilemma, vanilla contrastive learning is unable to model the correlations between temporally distant moments and learned inconsistent video representations. Both characteristics lead to vanilla contrastive learning being unsuitable for video grounding. In this paper, we introduce Geodesic and Game Localization (G2L), a semantically aligned and uniform video grounding framework via geodesic and game theory. We quantify the correlations among moments leveraging the geodesic distance that guides the model to learn the correct cross-modal representations. Furthermore, from the novel perspective of game theory, we propose semantic Shapley interaction based on geodesic distance sampling to learn fine-grained semantic alignment in similar moments. Experiments on three benchmarks demonstrate the effectiveness of our method. The code is available at \hyperlink{https://github.com/lihxxxxx/G2L}{https://github.com/lihxxxxx/G2L}.
\end{abstract}


\begin{figure}[t]
\begin{center}
   \includegraphics[width=1.0\linewidth]{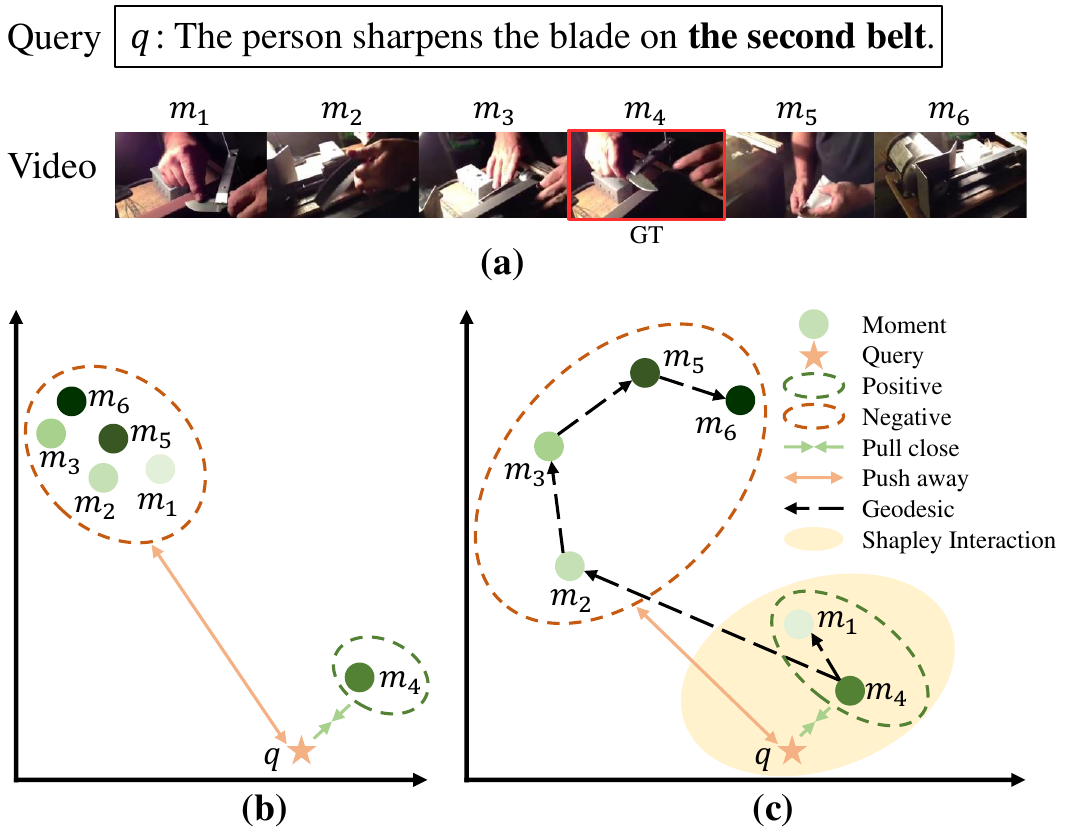}
\end{center}
\vspace{-0.3cm}
   \caption{(a) Illustration of video grounding. `GT' indicates the ground truth. A comparison of (b) existing contrastive learning-based methods and (c) our proposed G2L method. G2L makes semantically similar video moments closer in representation space while exploring nuances among similar moments.}
\label{fig:intro}
\end{figure}

\section{Introduction}
Video grounding~\cite{CTRLgao2017tall,MMNwang2022negative,SPLliu2022skimming,IVG-DCLnan2021interventional,Sscsding2021support,2D-TANzhang2020learning,cao2022deep,zhang2021cola,VSLNet_Lzhang2021natural,DRNzeng2020dense,cao2023iterative,jiang2022video} aims to identify the timestamps semantically corresponding to a given query within an untrimmed video, which is a challenging multimedia retrieval task due to flexibility and complexity of the query and video content. The video grounding model needs well to model the complex cross-modal correlations and semantic information. 


Contrastive learning~\cite{chen2020simple,he2020momentum,misra2020self} is proposed to learn representations by contrasting positive pairs against negative pairs. With the popularity of contrastive learning in vision-language tasks~\cite{li2021align,cheng2023ssvmr, radford2021learning,cao2022locvtp,cao2022correspondence,Yang_2023_ICCV, yang2023implicit}, several works also apply it to video grounding. 
Nan \etal~\cite{IVG-DCLnan2021interventional} propose a dual contrast learning to learn more informative feature representations by maximizing the mutual information between the query and the corresponding video clips. This naive solution, however, achieves sub-optimum performance.

Generally, contrastive learning requires two key properties~\cite{wang2020understanding}: \emph{alignment} and \emph{uniformity}. \emph{Alignment} favors encoders that assign similar features to similar samples. \emph{Uniformity} prefers a feature distribution that preserves maximal information, \ie, the uniform distribution on the unit hyper-sphere. We argue that these two issues are not satisfied in current video grounding works with contrastive learning.

Firstly, the semantic overlapping issue is widespread, \ie, the co-existence of some visual entities in both ground truth and other moments. As shown in Figure~\ref{fig:intro}(a), the entity of `\texttt{person}', `\texttt{blade}' and `\texttt{belt}' appear in both ground-truth moment $m_4$ and others. 
Since there exist no classification labels in video grounding, previous methods distinguish positive and negative samples only based on the annotated moments. 
This strict scheme, however, ignores the semantic overlapping among video moments, which leads to the contradiction in feature representations. 
As shown in Figure~\ref{fig:intro}(a)(b), the moment $m_1$ of ``\texttt{sharpen the blade on the first belt}'' only differs from target moment $m_4$ in the `\texttt{second}' order. They share similar semantic meanings but are forced to be pushed away in feature space. It is not consistent with the \emph{alignment} principle of the ideal contrastive learning. 

Another issue lies in the sparse annotation dilemma~\cite{DRNzeng2020dense, li2023generating}. Due to the costly labeling process, only a few moments are annotated regardless of the thousands of frames contained. Such severe data imbalance leads to significant learning bias for vanilla contrastive learning, \ie, unannotated moments are pushed away by different queries, regardless of the semantic relationships. Consequently, their representations are close, although they don't necessarily have strong semantic similarities.
This undermines the \emph{uniformity} requirement of contrastive learning. For example in Figure~\ref{fig:intro}(b), wrong results always exist when encountering these unannotated moments (\eg, $m_1$).

To address the issues mentioned above, we propose a novel Geodesic and Game Localization (G2L), a semantically aligned and uniform video grounding framework as shown in Figure~\ref{fig:intro}(c). We propose to measure the similarity according to the geodesic distance~\cite{kimmel1998computing} between two video moments along the manifold. In Figure~\ref{fig:intro}(c), the geodesic distance between $m_4$ to $m_6$ is the length of the shortest path as the moment graph, \ie $m_4\!\to\! m_2\!\to\!m_3\!\to\!m_5\!\to\!m_6$. 
In contrast to previous methods, we construct positive and negative pairs based on the geodesic distance rather than the temporal moment to relax the strict positional principle. The geodesics from the target moment to other moments are used to guide the maximizing mutual information. In this manner, the distance between video moments correctly reflects semantic relevance.
 
Unfortunately, the relaxed contrastive objective with geodesic leads to one side-effect, \ie, the model may confuse similar video moments. As shown in Figure~\ref{fig:intro}(c), the model may falsely map $m_1$ to be as close to the query $q$ since $m_1$ shares a similar appearance with the ground truth $m_4$. To further prevent the model from confusing similar video moments, 
we formulate video moments and queries as multiple players into a cooperative game and quantify their game-theoretic interactions (\ie, Shapley interactions~\cite{shapley1997value,grabisch1999axiomatic}). Through this, we evaluate the marginal contributions of each fine-grained component, which leads to a more accurate division. 
However, computing the exact Shapley interaction for all players is an NP-hard problem~\cite{matsui2001np} and is difficult to achieve solutions in the video grounding setting. Therefore, we further propose a semantic Shapley interaction module, which samples similar intra-video moments by geodesic distance with a focus on their nuances.

In sum, our contributions are summarized as follows:
\begin{itemize}[topsep=0pt, partopsep=0pt, leftmargin=13pt, parsep=0pt, itemsep=3pt]
    \item We present G2L which introduces geodesic and game theory to learn the semantic alignment and uniformity between video and query for video grounding.
    \item We propose a novel geodesic-guided contrastive learning scheme that considers the correct semantics of all moments in the video.
    \item We introduce an effective semantic Shapley interaction strategy based on geodesic distance.
    \item Extensive experiments on three public datasets demonstrate the effectiveness of our G2L.
\end{itemize}

\section{Related Work}
\noindent \textbf{Video Grounding.} Video grounding proposed by~\cite{CTRLgao2017tall,anne2017localizing}, which aims to predict the start and end boundaries of the activity described by a given language query within a video. Early approaches focus on carefully designed complex video-text interaction modules. Yuan \etal~\cite{yuan2019find} propose an approach which can directly predict the coordinates of the queried video clip using attention mechanism. Zeng \etal~\cite{DRNzeng2020dense} propose a pyramid neural network to consider multi-scale information. Liu \etal~\cite{liu2018attentive} advise a memory attention to emphasize the visual features and simultaneously utilize the context information. Xu \etal~\cite{xu2019multilevel} introduce a multi-level model to integrate visual and textual features earlier and further re-generate queries as an auxiliary task. To improve the representation of the model, several methods introduce contrastive learning or cross-modal discrimination. Ding \etal~\cite{Sscsding2021support} propose to combine discriminative contrastive objective and generative caption objective to optimize dual-encoder. Nan \etal~\cite{IVG-DCLnan2021interventional} introduce introducing causal intervention and dual contrastive learning to improve representation. In this paper, we propose to model semantic alignment and uniformity by approximate geodesics and game-theoretic interactions.

\noindent \textbf{Contrastive Learning.} Contrastive learning (CL) often serves as an unsupervised objective to learn representations by contrasting positive pairs against negative pairs~\cite{chen2020simple,he2020momentum,misra2020self,li2023unify,cheng_F,cheng2023acl}. Some prior works consider maximizing the mutual information (MI) between latent representations~\cite{hjelm2018learning}. MI quantifies the "amount of information" gained about one random variable by observing another random variable~\cite{bell1995information}. 
Contrastive learning has been applied to vision-language tasks to learn the joint representations of visual and textual modalities~\cite{miech2020end,sun2019learning}. Sun \etal~\cite{zhang2021video} propose a retrieval and localization network with contrastive learning for video corpus moment retrieval.

\noindent \textbf{Shapley Value.} The Shapley value~\cite{shapley1997value, jin2023video} originates from cooperative game theory. It has been theoretically proven to be the unique metric to fairly estimate the contribution of each player in a cooperative game such that satisfying certain desirable axioms~\cite{weber1988probabilistic} and is widely used in deep learning. Li \etal~\cite{li2022fine} propose a semantically aligned vision-language pre-training based on Shapley value to model fine-grained semantics. Ren \etal~\cite{ren2021unified} propose to explain adversarial attacks by Shapley value. Li \etal~\cite{li2021shapley} propose to explicit credit assignment for multi-agent reinforcement learning using Shapley value.

\begin{figure*}[t]
\begin{center}
   \includegraphics[width=0.95\linewidth]{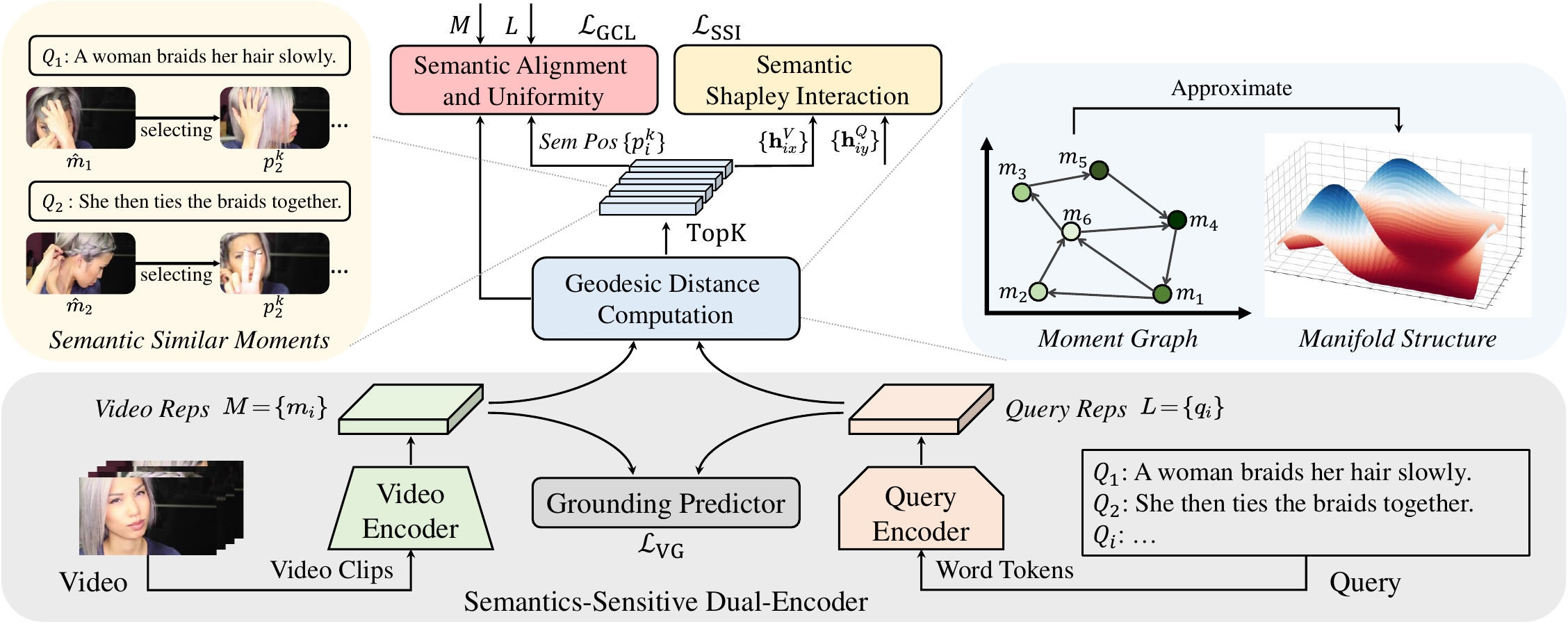}
\end{center}
   \vspace{-0.2cm}
   \caption{Overview of Geodesic and Game Localization (G2L). Our framework encourages the model to learn semantically aligned and uniform joint representations. In the inference stage, we directly fuse the video features and query features to compute the predicted moments. In the training stage, the grounding loss $\mathcal{L}_{\mathrm{VG}}$ is obtained by calculating the cross-entropy between the predicted moment and the target moment. Then, we approximate the high-dimensional manifold structure of the video representations through a moment graph and calculate the geodesic distance from the target moment to other moments. Finally, we leverage geodesic distance for cross-modal discrimination and semantic Sharpley interaction modeling.}
\label{fig:pipeline}
\end{figure*}

\section{Geodesic and Game Localization (G2L)}

\subsection{Problem Formulation and Model Overview}
Let $Q_i$ and $V_i$ be a given textual query and untrimmed video respectively. The purpose of video grounding is to locate the most relevant video interval $A_i=(t_i^s, t_i^e)$, where $t_i^s$ and $t_i^e$ are starting and ending times respectively. The key to video grounding is to learn semantics between video and query. To this end, previous methods~\cite{IVG-DCLnan2021interventional, Sscsding2021support} incorporate vanilla contrastive learning into existing cross-modal interaction architectures. Typically, at the training stage, they employ two loss functions: video grounding loss $\mathcal{L}_{\mathrm{VG}}$ and vanilla contrastive loss $\mathcal{L}_{\mathrm{VCL}}$. $\mathcal{L}_{\mathrm{VG}}$ computes the cross entry between the target timestamp and the prediction timestamp to optimize the model. $\mathcal{L}_{\mathrm{VCL}}$ discriminates between positive samples and negative samples based on the temporal moment and adopts noise-contrastive estimation (NCE)~\cite{gutmann2010noise} to obtain MI of videos and queries. 

However, vanilla contrastive learning is not appropriate for video grounding due to the semantic overlapping and sparse annotation dilemma. To learn correct semantics and improve representations, as illustrated in Figure~\ref{fig:pipeline}, we propose Geodesic and Game Localization (G2L), a semantically aligned and uniform video grounding framework that germinates from geodesic and cooperative game theory. Our G2L learns semantic alignment and uniformity from two components. With geodesic-guided contrastive learning (GCL), complete semantic alignment and uniformly distributed video features are learned. By semantic Shapley interaction(SSI), we learn the fine-grained semantic alignment between similar moments and target queries. Combined with the two novel proposed training objectives, the full training objective of semantically aligned and uniform video grounding can be formulated as:
\begin{equation}
    \mathcal{L} = \mathcal{L}_{\mathrm{VG}}+ \mathcal{L}_{\mathrm{GCL}} + \mathcal{L}_{\mathrm{SSI}} 
\end{equation}
During inference, it can be directly removed, rendering a semantics-sensitive dual encoder.

\subsection{Feature Encoder}

\noindent \textbf{Video Encoder.} For an input video $V_i$, we first segment it into small video clips and perform a fixed-interval sampling over these clips. Then the video clips are fed into a pre-trained 3D CNN model (\eg C3D) to extract the video features $F^V_i$. Referring to the previous works~\cite{2D-TANzhang2020learning}, we employ sparse sampling and a proposal network to construct the feature map $F^M_i\!=\!\{m_i\}^{N_m}_{i=1}$ of moment candidates based on $F^V_i$, where $N_m$ is the number of proposals in the video.

\noindent \textbf{Query Encoder.} For a textual 
 query $Q_i$, we generate the tokens of words by the tokenizer and add a class embedding token `[CLS]' in the beginning. Then the tokens are fed into pre-trained BERT~\cite{kenton2019bert}, and we perform average pooling of its last two hidden states to obtain the sentence feature $q_i$.

\subsection{Geodesic Distance Computation}
Video content is intricate and flexible,  and for video clips, temporal proximity does not always equate to semantic similarity. Conversely, two video clips can have strong correlations even though they are temporally distant. We propose to measure correlations of video representations using the geodesic distance. Formally, we define a mini-batch of video-query pairs as $\{V_i, Q_i\}_{i=1}^B$, where $B$ is the size of the mini-batch. After fed into the video and query encoders, we obtain textual representation $L\!=\!\{q_i\}_{i=1}^B$ and moment representations $M\!=\!\{m_i\}_{i=1}^{B\times N_m}$.

The video moments representations may lie in a high-dimensional manifold, and our purpose is to measure the geodesic distance between two points along the manifold. However, computing the exact geodesic distance~\cite{kimmel1998computing} is difficult without explicit knowledge of the manifold structure. To obtain the geodesic distance, we first employ the K-NN graph~\cite{cover1967nearest} to approximate the manifold structure~\cite{surazhsky2005fast, chowdhury2022unsupervised}. In this graph, each moment $m_i$ forms a node, and each node connects to at most $n$ other nodes. A directed edge exists from node $m_i$ to node $m_j$ if node $m_j$ is among the $n$ nearest neighbors of $m_i$. The edge weight $d(m_i, m_j)$ is defined using cosine similarity: $d(m_i, m_j) \!=\! 1 - m_im_j^\top$. 
Finally, we utilize the shortest path algorithm, \ie, Djikstra~\cite{dijkstra2022note}, to compute the length of the shortest path between two moments along the obtained weighted directed graph as the geodesic distance $\mathcal{G}(m_i, m_j)$.

\subsection{Geodesic-guided Contrastive Learning}

\noindent \textbf{Previous Contrastive Scheme.} 
In video grounding, a query usually corresponds to multiple clips. An intuitive method~\cite{Sscsding2021support} to learn representations is to set the clips in ground truth intervals as positive samples, while others are negatives. Another scheme~\cite{MMNwang2022negative} is to calculate the intersection over union (IoU) between the ground truth and other moments, where those with higher IoU are considered as positive samples and the lower ones are regarded as negative samples. Then they learn the joint representation by pulling the query features and ground truth moment features together, and pushing the query features and non-ground truth moment features apart. The previous contrastive loss $\mathcal{L}_{\mathrm{VCL}}$ can be formulated as:
\begin{equation}
    \mathcal{L}_{\mathrm{VCL}}=-\sum_{i=1}^{B}\left(\log\frac{\mathrm{exp}(q_im_i^\top/\tau)}{\sum\limits_{m_j \in M}\mathrm{exp}(q_im_j^\top/\tau) }\right) 
    \label{eq: ocl}
\end{equation}
where $\tau$ is the temperature hyper-parameter. 

While intuitive, such a manner fails to capture semantic \emph{alignment} and \emph{uniformity} between queries and video content. To learn complete semantic \emph{alignment} and \emph{uniformity} for video grounding, we propose geodesic-guided contrastive learning.

\noindent \textbf{Semantic Alignment.} In vanilla supervised contrastive learning~\cite{wang2022partial, khosla2020supervised}, all classes play the same role: pulling is done within every class and pushing between every pair of different classes. Such uniform contrastive learning works well for symmetric and equal cross-modal learning. 
In fact, due to semantic overlapping, the semantics of the video and text are asymmetric and unequal in video grounding. Temporally distant moments may have similar semantics. To model semantic \emph{alignment}, we discriminate positive pairs based on the geodesic distance instead of the temporal moment. For a query feature $q_i$, we define the $k$ moments with the closest geodesic distance from the target moment $\hat{m}_i\!\in\! M$ as its semantic positive samples $P_i$:
\begin{equation}
P_i = \left\{p_i^k\right\} =\underset{k}{\arg \mathrm{topk} \mathcal{G}\left(\hat{m}_i\right.}, m_k)
\label{eq:p_i}
\end{equation}
$P_i$ contains $m_i$ and moments adjacent to $m_i$ geodesic which is used to construct positive pairs with $q_i$ to relax the previous  strict positional principle.

\noindent \textbf{Semantic Uniformity.} Due to the sparse annotation dilemma~\cite{DRNzeng2020dense, li2023generating}, most of the moments are only marked as negative while pushed away by different queries. Notably, this push operation is undifferentiated, \ie, the model is encouraged to learn to push all unannotated moments from the annotated ones. This clearly undermines the model to learn correct semantics. To learn semantic \emph{uniformity}, we introduce the geodesic distance to differentially push negative samples away from the query based on semantic relationships. The similarity between query and moment $s(q_i,m_j)$ is defined as:

\begin{equation}
\!s(q_i,m_j) \!= \!\mathrm{exp}\!\left(\!q_i m_j^\top
\!\left(\hat{m}_im_j^\top \mathrm{log}\frac{1}{\mathrm{exp}\!\left(\mathcal{G}\left(\hat{m}_i,m_j\right) \!+\!1 \right)}\!   \right)\right)\!
\end{equation}
We assign corresponding weights according to the geodesic distance $\mathcal{G}\left(\hat{m}_i,m_j\right)$ between the target moment $\hat{m}_i$ and all moments $m_j$. $s(q_i,m_j)$ considers the relationships between negative samples while maximizing mutual information.

Finally, our geodesic-guided contrastive loss can be formulated as:
\begin{equation}
    \mathcal{L}_{\mathrm{GCL}}=-\sum_{i=1}^{B}\left(\log\frac{\sum\limits_{p_i^k \in P_i} \mathrm{exp}(q_i{p_i^k}^\top/\tau)}{\sum\limits_{m_j \in M} s(q_i, m_j)/\tau}\right) 
    \label{eq: gcl}
\end{equation}
In contrast to Equation~\ref{eq: ocl}, $\mathcal{L}_{\mathrm{GCL}}$ is designed for representation learning rather than directly learning localization.

\subsection{Semantic Shapley Interaction Modeling}
To prevent the model from confusing similar video moments due to the relaxed contrastive objective in Equation~\ref{eq:p_i}, we propose semantic Shapley interaction to model fine-grained semantic alignment.
\vspace{-0.3cm}
\subsubsection{Preliminaries}
\noindent \textbf{Shapley Value.}\, The Shapley value~\cite{shapley1997value} is a classical game theory solution for the unbiased estimation of the contribution of each player in a cooperation game. Assume a game consists of $\mathcal{N}$ players, $\mathcal{U} \subseteq \mathcal{N}$ represents a potential subset of players. A game function $f(\cdot)$ maps each subset $\mathcal{U}$ of players to a score, estimating the cooperated contribution of a set of players. For a player $i$, its Shapley value $\phi(i|\mathcal{N})$ is computed as the average marginal contribution of player $i$ to all possible coalitions $\mathcal{U}$ without $i$:
\begin{gather}
    \phi(i \mid \mathcal{N})=\sum_{\mathcal{U} \subseteq \mathcal{N} \backslash\{i\}} p(\mathcal{U})[f(\mathcal{U} \cup\{i\})-f(\mathcal{U})] \label{eq: Shapley value}
    \\ \quad p(\mathcal{U})=\frac{|\mathcal{U}| !(|\mathcal{N}|-|\mathcal{U}|-1) !}{|\mathcal{N}| !} 
\end{gather}
where $p(\mathcal{U})$ is the likelihood of $\mathcal{U}$ being sampled. The Shapley value has been proved to be the unique metric that satisfies the following axioms: \emph{Linearity}, \emph{Symmetry}, \emph{Dummy}, and \emph{Efficiency}. We summarize these axioms in the supplementary material.

\noindent \textbf{Shapley Interaction.}\, In game theory, some players tend to form a small cooperative coalition and always participate in the game together. This cooperation provides additional contributions to the game. The Shapley interaction~\cite{grabisch1999axiomatic} measures the additional contributions made by the coalition compared to players working individually. We define $[\mathcal{U}]$ as a single hypothetical player, which is the union of the players in $\mathcal{U}$. A reduced game is formed by removing the individual players in $\mathcal{U}$ from the game and adding $[\mathcal{U}]$ to the game. Finally, according to Equation~\ref{eq: Shapley value}, the Shapley interaction for coalition $\mathcal{U}$ is formulated as:
\begin{equation}
    \!\mathfrak{I}([\mathcal{U}])=\phi([\mathcal{U}] \mid \mathcal{N} \backslash \mathcal{U} \cup\{[\mathcal{U}]\}) - \sum_{i \in \mathcal{U}} \phi(i \mid \mathcal{N} \backslash \mathcal{U} \cup\{i\})\!
    \label{eq: Shapley interaction}
\end{equation}
where $\mathfrak{I}([\mathcal{U}])$ reflects the interactions inside $\mathcal{U}$. The higher value of $\mathfrak{I}([\mathcal{U}])$ indicates that players in $\mathcal{U}$ cooperate closely with each other.

\subsubsection{Fine-Grained Semantic Alignment via Shapley}
\label{sec: SSI}

\quad According to Equation~\ref{eq: Shapley value} and Equation~\ref{eq: Shapley interaction}, it can be known that the computational complexity of the Shapley interaction grows factorially as the number of players increases. To reduce the computational cost, we propose semantic Shapley interaction based on geodesic distance sampling. Specifically, as in Equation~\ref{eq:p_i}, we sample semantic similar moments based on geodesic distances to investigate their nuances. Then, we consider queries and semantic similar moments from the same video as players in the same cooperative game.

To avoid confusion with the previous symbols, we define $\mathcal{H}^V_i=\{\mathbf{h}^V_{ix}\}_{x=1}^{N^v_i}$ and $\mathcal{H}^Q_i=\{\mathbf{h}^Q_{iy}\}_{y=1}^{N^q_i}$ are a set of semantic positive samples and queries from the video $V_i$ respectively, where $N_i^v \!=\!K N_i^q $ and $N_i^q$ indicate the number of similar moments and queries in video $V_i$ respectively. $K$ is the number of moments sampled for each query $\mathbf{h}^Q_{iy}$. We investigate the effect of $K$ on computational cost and performance in the supplementary material.

If a query and a video moment have strong semantic correspondence, then they tend to cooperate with each other and contribute to the coalition. Thus, we can take $\mathcal{H}^i\!=\!\mathcal{H}_i^V \!\cup\! \mathcal{H}_i^Q$ as the players in the same game inspired by~\cite{li2022fine}. 
We define the alignment matrix as: $\mathcal{A}_i\!=\![a^i_{xy}]$, where $a^i_{xy}\!=\!{h^V_{ix}}^\top h^Q_{iy}$ represents the alignment score between $x$-th moment and $y$-th query in video $V_i$. Next, $\tilde{\mathcal{A}}_i$ is obtained by applying softmax-normalization over each row of $\mathcal{A}_i$. Then we average the maximum alignment score $\mathrm{max}_y\tilde{a}^i_{xy}$ as the fine-grained moment-to-query similarity $\psi_1$. Similarly, we can obtain the fine-grained query-to-moment similarity $\psi_2$. The total fine-grained similarity score can be defined: $\psi=(\psi_1+\psi_2)/2$, which be considered as the game score $f(\cdot)$ in our game.

According to Equation~\ref{eq: Shapley interaction}, the semantic Shapley interaction of them can be formulated as:
\begin{align}
\!\mathfrak{I}\big(\left[\mathcal{H}^i_{x y}\right]\big)
\!&\!=\phi\big(\left[\mathcal{H}_{x y}^i\right] \mid \mathcal{H}^i \backslash \mathcal{H}_{x y}^i \cup\left\{\left[\mathcal{H}_{x y}^i\right]\right\}\big) \nonumber \\
\!&-\phi\big(\mathbf{h}_{ix}^{V} \mid \mathcal{H}^i \backslash \mathcal{H}_{x y}^i \cup\big\{\mathbf{h}_{ix}^{V}\big\}\big) \nonumber \\
\!&-\phi\big(\mathbf{h}_{iy}^{Q} \mid \mathcal{H}^i \backslash \mathcal{H}_{x y}^i \cup\big\{\mathbf{h}_{iy}^{Q}\big\}\big) \\
\!&\!=\underset{\mathcal{C}}{\mathbb{E}}\big\{{\substack{\mathcal{U} \subseteq \mathcal{H}^i \backslash \mathcal{H}^i_{x y} \\|\mathcal{U}|=\mathcal{C}}}{\mathbb{E}}[f\big(\mathcal{U} \cup \mathcal{H}^i_{ x y}\big) \!-\!f\big(\mathcal{U} \cup\left\{\mathbf{h}_{ix}^{V}\right\}\big) \nonumber \\
\!&-f\big(\mathcal{U} \cup\big\{\mathbf{h}_{iy}^{Q}\big\}\big)+f(\mathcal{U}) ] \big\}\!
\label{eq: ssi}
\end{align}
where $\mathfrak{I}\left(\left[\mathcal{H}^i_{ x y}\right]\right)$ represents the single player formed by the coalition of $x$-th moment and $y$-th query in video $V_i$, $\mathcal{C}$ represents the coalition size. We take normalized $\mathfrak{I}^{\prime}\left(\left[\mathcal{H}^i_{x y}\right]\right)$ as soft labels, the fine-grained semantic alignment loss can be defined as: 
\begin{equation}
    \mathcal{L}_{\mathrm{SSI}}=-\sum_{i=1}^{T} \frac{1}{N_i^v N_i^q} 
    \sum_{x=1}^{N_i^{v}} \sum_{y=1}^{N_i^{q}}
    \mathfrak{I}^{\prime}\left(\left[\mathcal{H}^{i}_{x y}\right]\right) \log \left(\tilde{a}^{i}_{x y}\right)
\end{equation}
where $T$ is the total number of unique videos in a mini-batch, \ie $T\le B$. 
\section{Experiments}

\subsection{Datasets and Evaluation}

\textbf{ActivityNet-Captions.} ActivityNet-Captions~\cite{krishna2017dense} contains 20,000 untrimmed videos and 100,000 descriptions~\cite{caba2015activitynet}, covering a wide range of complex human behavior. The video clips with annotations have much larger variations. Following the public split~\cite{2D-TANzhang2020learning}, we use 37417, 17505 and 17031 sentence-video pairs for training, validation and testing, respectively.

\textbf{Charades-STA.} The Charades dataset~\cite{sigurdsson2016hollywood} is collected for video action recognition and video captioning. Gao \etal~\cite{CTRLgao2017tall} adapt the Charades dataset to the video grounding task by collecting the query annotations. The Charades-STA dataset contains 6672 videos and involves 16128 video-query pairs, where 12408 pairs are used for training and 3720 for testing. We follow the same split of the dataset as in Gao \etal~\cite{CTRLgao2017tall} for fair comparisons.

\textbf{TACoS.} TACoS~\cite{regneri2013grounding} contains 127 videos from the cooking scenarios. 
We follow the standard split ~\cite{CTRLgao2017tall}, which has 10146, 4589 and 4083 video query pairs for training, validation and testing, respectively.


\textbf{Evaluation.} Following previous work \cite{CTRLgao2017tall,2D-TANzhang2020learning}, we adopt ``R@n, IoU=m'' as the evaluation metric. It calculates the percentage of IoU greater than ``m'' between at least one of the top ``n'' video moments retrieved and the ground truth. 

\begin{table}[t]
\renewcommand\arraystretch{1.1}
\begin{center}
\begin{tabular}{c|cccc}
\hline
\multirow{2}{*}{Model} & R@1 & R@1 & R@5 & R@5 \\
 & IoU0.5 & IoU0.7 & IoU0.5 & IoU0.7 \\ \hline\hline
CTRL~\cite{CTRLgao2017tall} & 29.01 & 10.34 & 59.17 & 37.54 \\
ACRN~\cite{liu2018attentive} & 31.67 & 11.25 & 60.34 & 38.57 \\
TripNet~\cite{hahn2019tripping} & 32.19 & 13.93 & - & - \\
SCDM~\cite{SCDMyuan2019semantic} & 36.75 & 19.86 & 64.99 & 41.53 \\
LGI~\cite{LGImun2020local} & 41.51 & 23.07& &  \\
VSLNet~\cite{VSLNetzhang2020span} & 43.22 & 26.16 & - & -\\
2D-TAN~\cite{2D-TANzhang2020learning} & 44.51 & 26.54 & 77.13 & 61.96 \\
DRN~\cite{DRNzeng2020dense} & 45.45 & 24.39 & 77.97 & 50.30 \\
DPIN~\cite{DPINwang2020dual} & 47.27 & 28.31 & 77.45 & 60.03  \\ 
CBLN~\cite{CBLNliu2021context} & 48.12 & 27.60 & 79.32 & 63.41 \\
SMIN~\cite{SMINwang2021structured} & 48.46 & 30.34 & \underline{81.16} & 62.11 \\
MATN~\cite{MATNzhang2021multi} & 48.02 & \underline{31.78} & 78.02 & 63.18 \\
GTR~\cite{GTRcao2021pursuit} & \underline{50.57} & 29.11 & 80.43 & \underline{65.14} \\
CMAS~\cite{yang2022video} &46.23 & 29.48 & 77.04 & 60.25 \\
MMN~\cite{MMNwang2022negative} & 48.59 & 29.26 & 79.50 & 64.76 \\
\hline
\emph{\textbf{CL-based:}} &  &  &  &  \\
IVG-DCL~\cite{IVG-DCLnan2021interventional} & 43.84 & 27.10 & - & - \\
SSCS~\cite{Sscsding2021support} &  46.67 & 27.56 & 78.37 & 63.78 \\
\hline      
G2L (ours) & \textbf{51.68} & \textbf{33.35} & \textbf{81.32} & \textbf{67.60} \\ \hline
\end{tabular}
\end{center}
\caption{Performance comparisons on ActivityNet-Captions using C3D features.}
\label{tab:anet}
\vspace{-1.em}
\end{table}

\subsection{Implementation Details}
For a fair comparison, we extracted video features from a pre-trained 3D CNN (C3D for ActivityNet-Captions and TACoS, VGG for Charades-STA) following previous works~\cite{MMNwang2022negative,Sscsding2021support,IVG-DCLnan2021interventional}. We uniformly sampled (256, 32, 256) clips as the input video sequence and set the length of 2D feature map~\cite{2D-TANzhang2020learning} (64, 16, 128) for ActivityNet-Captions, Charades-STA, and TACoS, respectively. For the language query, the pre-trained BERT\cite{kenton2019bert} was employed for each word of the query. The average pooling outputs of the first and last layers were used to obtain the embedding of the whole sentence. We employed the AdamW optimizer~\cite{loshchilov2018decoupled} to train our model and set the temperature weight $\tau$ to 0.1. The learning rates were set to ($8 \times 10^{-4}$, $8 \times 10^{-4}$, $1 \times 10^{-4}$) for ActivityNet-Captions, Charades-STA, and TACoS, respectively. We conducted experiments on 8 A100 GPUs with batch size 48 for ActivityNet-Captions and Charades-STA, and on 4 A100 GPUs with batch size 8 for TACoS. 

\begin{table}[t]
\renewcommand\arraystretch{1.1}
\begin{center}
\begin{tabular}{c|cccc}
\hline
\multirow{2}{*}{Model} & R@1 & R@1 & R@5 & R@5 \\
 & IoU0.5 & IoU0.7 & IoU0.5 & IoU0.7 \\ \hline\hline
MCN~\cite{anne2017localizing} & 17.46 & 8.01 & 48.22 & 26.73 \\
SAP~\cite{chen2019semantic} & 27.42 & 13.36 & 66.37 & 38.15 \\
TripNet~\cite{hahn2019tripping} & 36.61 & 14.50 - & - \\
MAN~\cite{zhang2019man} & 41.21 & 20.54 & 83.21 & 51.85 \\
2D-TAN~\cite{2D-TANzhang2020learning} & 39.70 & 23.31 & 80.32 & 51.26 \\
DRN~\cite{DRNzeng2020dense} & 42.90 & 23.68 & 87.80 & 54.87 \\
FVMR~\cite{FVMRgao2021fast} & 42.36 & 24.14 & 83.97 & 50.15 \\
CBLN~\cite{CBLNliu2021context} & 43.67 & 24.44 & 88.39 & 56.49 \\
MMRG~\cite{zeng2021multi} & 44.25 & - & 60.22 & - \\
MMN~\cite{MMNwang2022negative} & \underline{47.31} & \underline{27.28} & \underline{83.74} & \underline{58.41} \\
\hline
\emph{\textbf{CL-based:}} &  &  &  &  \\
SSCS~\cite{Sscsding2021support} & 43.15 & 25.54 & 84.26 & 54.17 \\
\hline
\textbf{G2L(Ours)} & \textbf{47.91} & \textbf{28.42} & \textbf{84.80} & \textbf{59.33} \\ \hline
\end{tabular}
\end{center}
\caption{Performance comparisons on Charades-STA using VGG features.}
\label{tab:charades}
\vspace{-1.em}
\end{table}

\subsection{Comparisons with State-of-the-art Methods}

\textbf{Comparison on ActivityNet-Captions.} In Table~\ref{tab:anet}, we report our performance in comparison with other state-of-the-art methods on ActivityNet-Captions. Compared with IVG-DCL~\cite{IVG-DCLnan2021interventional} and SSCS~\cite{Sscsding2021support}, which are also based on contrastive learning, our method achieves significant improvements. Notably, ActivityNet-Captions is currently the dataset with the most severe semantic overlapping and sparse annotation dilemma. We observe that our model showed even more significant improvements on ActivityNet-Captions, achieving absolute improvements of up to 7.8\% and 5.7\% compared to IVG-DCL~\cite{IVG-DCLnan2021interventional} and SSCS~\cite{Sscsding2021support}, respectively. They encourage the model to focus on cross-modal alignment in favor of grounding while ignoring the semantics among full moments, especially unannotated moments. In contrast, our method conducts cross-modal discrimination guided by the geodesic distance such that overcoming data bias. 

\textbf{Comparison on Charades-STA.} Table~\ref{tab:charades} reports the result comparison between state-of-the-art methods on Charades-STA. Compared with SSCS~\cite{Sscsding2021support}, our method achieves a performance improvement of up to 5.1\%. On more stringent evaluation metrics, such as ``R@1 IoU=0.7'', our method achieves a performance improvement of 1.1\% compared to the cutting-edge method MMN~\cite{MMNwang2022negative}, which indicates that exploring fine-grained semantic alignment information between similar video moments can improve grounding quality. Notably, MMN~\cite{MMNwang2022negative} and SSCS~\cite{Sscsding2021support} have similar loss functions, and MMN~\cite{MMNwang2022negative} from a perspective on temporal grounding as a metric-learning problem proposes a mutual matching network to enhance joint representation learning by mining more negative samples achieving better performances. However, it also constructs cross-modal pairs based on the temporal moment, which results in a large number of constructed negative pairs containing potential weak semantic positive pairs, thus hindering learning.

\textbf{Comparison on TACoS.} Table~\ref{tab:tacos} summarizes the comparisons on the TACoS. We observe that our model achieves state-of-the-art results in most settings. However, the performance gain on this dataset is smaller than the previous three datasets. The reason is that the sparse annotation dilemma and semantic overlapping are insignificant on TACoS. It contains only 127 videos but has about 20,000 queries and focuses on cooking activities with more uniform objects, roles, and actions.
Nevertheless, our method still outperforms previous contrastive learning-based methods in various metrics.

\begin{table}[t]
\begin{center}
\begin{tabular}{c|cccc}
\hline
\multirow{2}{*}{Model} & R@1 & R@1 & R@5 & R@5 \\
 & IoU0.3 & IoU0.5 & IoU0.3 & IoU0.5 \\ \hline \hline
CTRL~\cite{CTRLgao2017tall} & 18.32 & 13.30 & 36.69 & 25.42 \\
CBP~\cite{CBPwang2020temporally} & 27.31 & 24.79 & 43.64 & 37.40 \\
SCDM~\cite{SCDMyuan2019semantic} & 26.11 & 21.17 & 40.16 & 32.18 \\
2D-TAN~\cite{2D-TANzhang2020learning} & 37.29 & 25.32 & 57.81 & 45.04 \\
DRN~\cite{DRNzeng2020dense} & -- & 23.17 & -- & 33.36 \\
CMIN~\cite{CMINzhang2019cross} & 24.64 & 18.05 & - & - \\
CSMGAN~\cite{CSMGANliu2020jointly} & 33.90 & 27.09 & 53.98 & 41.22 \\
CBLN~\cite{CBLNliu2021context} & 38.98 & 27.65 & 59.96 & 46.24 \\
MMN~\cite{MMNwang2022negative} & 39.24 & 26.17 & 62.03 & 47.39 \\
GTR~\cite{GTRcao2021pursuit} & 40.39 & \underline{30.22} & 61.94 & 47.73 \\
FVMR~\cite{FVMRgao2021fast} & \underline{41.48} & 29.12 & \underline{64.53} & \textbf{50.00} \\
\hline
\emph{\textbf{CL-based:}} &  &  &  &  \\
IVG-DCL~\cite{IVG-DCLnan2021interventional} & 38.84 & 29.07 & - & - \\
SSCS~\cite{Sscsding2021support} & 41.33 & 29.56 & 60.65 & 48.01 \\
\hline
\textbf{G2L(Ours)} & \textbf{42.74} & \textbf{30.95} & \textbf{65.83} & \underline{49.86} \\ \hline
\end{tabular}
\end{center}
\caption{Performance comparisons on TACoS using C3D features.}
\label{tab:tacos}
\vspace{-1.em}
\end{table}

\begin{table}[t]
\renewcommand\arraystretch{1.1}
\begin{center}
\begin{tabular}{l|cccc}
\hline
\multirow{2}{*}{Model} & R@1 & R@1 & R@5 & R@5 \\ 
& IoU0.5 & IoU0.7 & IoU0.5 & IoU0.7 \\ \hline \hline
Full Model & 51.68 & 33.35 & 81.32 & 67.60  \\
w/o SA & 50.85& 30.73  & 79.22 & 65.47 \\
w/o SU & 49.10 & 31.14 &  79.48 & 65.46 \\ 
w/o GCL & 48.74 & 28.62 & 78.89  & 64.44  \\
w/o SSI & 50.65 & 30.01 & 79.54 & 66.76 \\
w/o GCL+SSI & 45.74 & 26.74 & 77.26 & 61.64 \\ \hline
\end{tabular}
\end{center}
\caption{Ablation studies of main components on ActivityNet-Captions. ``SA" and ``SU" denote the Semantic Alignment and the Semantic Uniformity in GCL, respectively.}
\label{tab:abs component}
\end{table}

\section{Ablation Study}
\textbf{Effectiveness of Individual Components.}
In Table~\ref{tab:abs component}, we conduct a thorough ablation study on the proposed components to verify their effectiveness.
As shown in Table~\ref{tab:abs component}, removing the entire GCL will result in up to 4.5\% of performance degradation, demonstrating the contribution of GCL by learning semantic alignment and unification.  We also observe that removing either SA or SU results in about 2 points of performance drop on average, indicating that our method enables the vanilla contrastive learning to be more suitable for video grounding setting. Removing SSI leads to a 3.3\% drop in performance on the more stringent metric (\ie, ``R@1 IoU=0.7"), which highlights the importance of fine-grained semantic alignment for high-quality moment retrieval. The last row shows our baseline, and our method achieves up to 6 points of performance improvement without modifying the architecture compared to the baseline, confirming the superiority of our model.

\begin{table}[t]
\begin{center}
\begin{tabular}{c|cccc}
\hline
\multirow{2}{*}{Metric} & R@1 & R@1 & R@5 & R@5 \\ 
 & IoU0.5 & IoU0.7 & IoU0.5 & IoU0.7 \\ \hline
Euclidean & 47.83 & 28.44 & 79.42 & 65.46 \\
Timestamp & 49.85 & 29.73  & 78.15 & 65.47 \\
Cosine & 50.94 & 30.09 & 79.57 & 65.84 \\ \hline
\textbf{Geodesic }& \textbf{51.68} & \textbf{33.35} & \textbf{81.32 }& \textbf{67.60}  \\ \hline
\end{tabular}
\end{center}
\caption{Comparison of different distance metrics on ActivityNet-Captions.}
\label{tab:abs dis}
\vspace{-1.em}
\end{table}

\textbf{Effectiveness of Geodesic Distance.}
To demonstrate the effectiveness of the geodesic, we substitute it with different distance metrics in the model, including euclidean distance, timestamp distance, and cosine distance, and compare their performances as shown in Table~\ref{tab:abs dis}. It can be observed that when using euclidean distance, the performances drop by up to 5\%. This is due to the fact that video representations often lie in a high dimensional manifold and euclidean distance cannot accurately quantify the correlations among video moments. When using timestamp distance, the performances decrease by 2\% on average, indicating that the temporal adjacency does not necessarily correspond to semantic similarity due to the flexibility and complexity of video content. Using cosine distance results in performances dropping by up to 3\%. Due to the insufficient representation capacity of the model in the early training stage to accurately calculate similarity. We approximate the manifold structure of video features using the moment graph, where the reachability and the shortest path in the graph can facilitate similarity measurement.

\textbf{The efficiency of our method.}
In Table~\ref{tab: cost}, we compute the average training time per iteration and total inference time. 
Due to the K-NN graph and Shapley interaction, G2L requires more training costs. 
During inference, GCL and SSI can be removed thus our G2L only needs additional \textbf{3s} compared to the base model.

\begin{table}[t]
\begin{center}
\begin{tabular}{lcc}
\hline
Model & Iteration Time & Inference Time\\
\hline
2D-TAN~\cite{2D-TANzhang2020learning} & 0.13s & 32s \\
MMN~\cite{MMNwang2022negative} & 0.32s &  37s \\
\hline
Base Model & 0.39s & 40s \\
\textbf{\textbf{G2L(Ours)}} & 0.84s & 43s \\
\hline
\end{tabular}
\end{center}
\caption{Time consumption
on ActivityNet-Captions.}
\label{tab: cost}
\end{table}

\section{Qualitative Analysis}
\begin{figure}[t]
\begin{center}
   \includegraphics[width=1.0\linewidth]{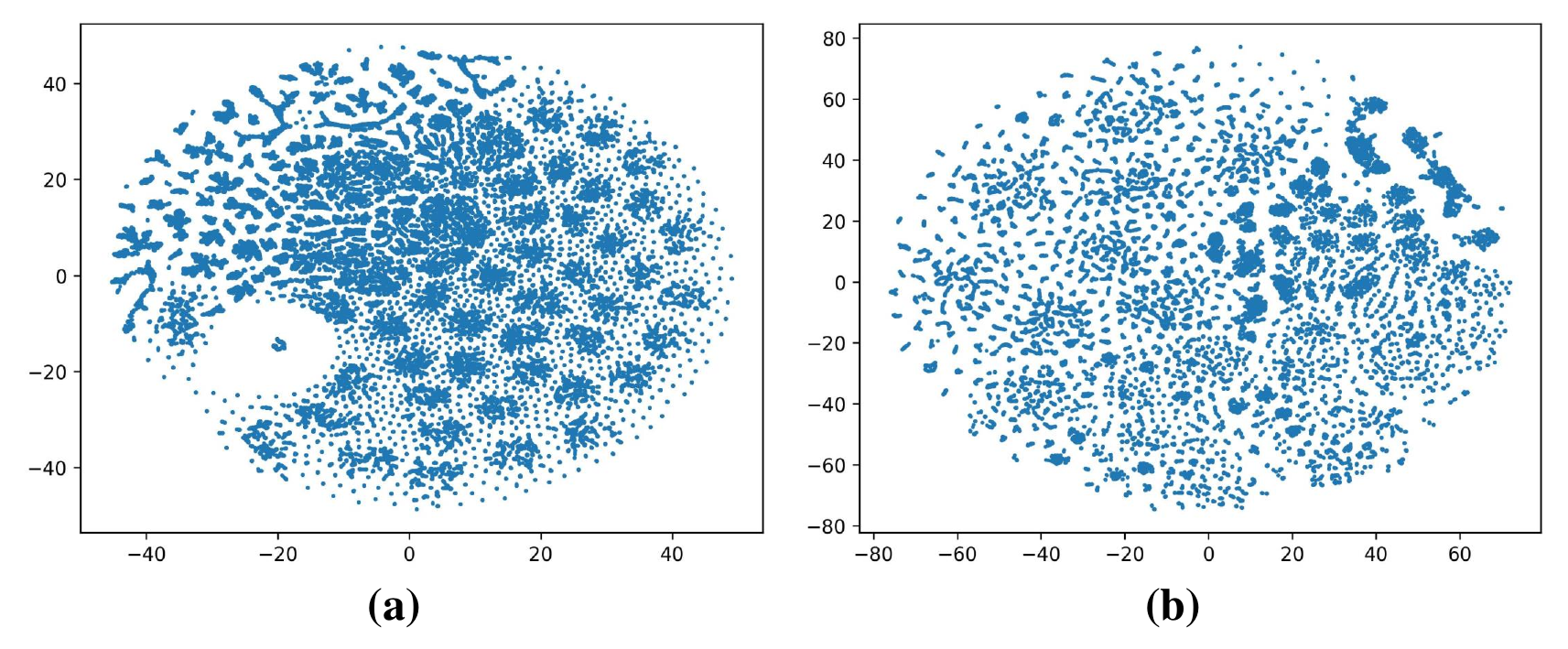}
\end{center}
\vspace{-0.3cm}
\caption{Projected video moment features (a): learned representations of the previous method with vanilla contrastive learning; (b): learned representations of our method.}
\vspace{-0.25cm}
\label{fig:vis2}
\end{figure}

\begin{figure}[t]
\begin{center}
   \includegraphics[width=0.95\linewidth]{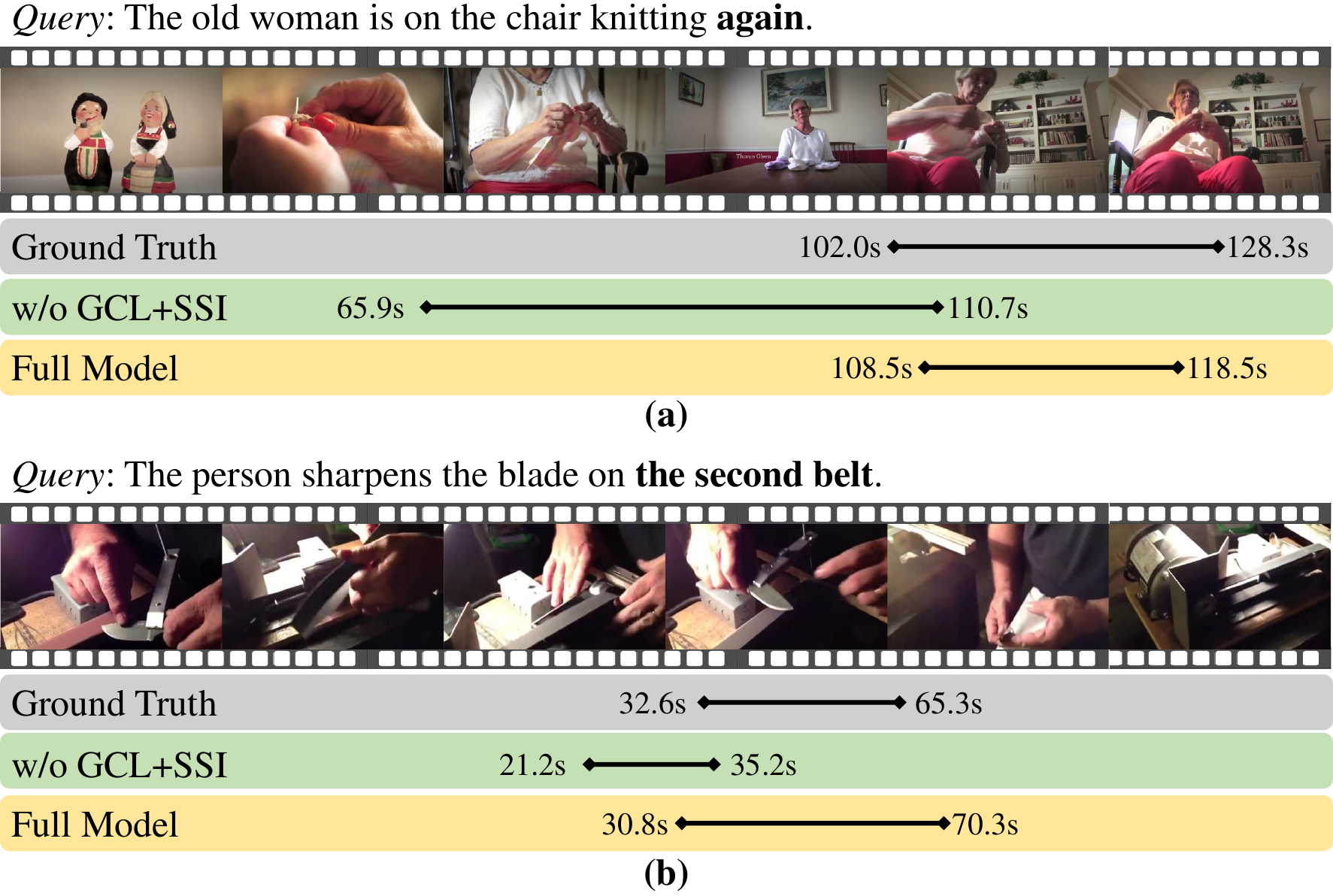}
\end{center}
\vspace{-0.3cm}
\caption{Qualitative results of our method on the ActivityNet-Captions.}
\vspace{-0.25cm}
\label{fig:vis1}
\end{figure}

We visualize a sample video from the ActivityNet-Captions into the 2D images using t-SNE~\cite{van2008visualizing}. In Figure~\ref{fig:vis2}(a), we observe that there is a clear ``island" in the learned representations of the previous method. We argue that it is formed by a few moments which are temporally adjacent to the ground truth. The reason is that the previous method divides positive and negative samples based on the strict temporal moment while pushing negative samples away by different queries, regardless of the semantic relationships.
In contrast, our method relaxes the previous strict contrastive objective using geodesic distance, mitigating semantic overlapping and sparse annotation dilemma. Furthermore, the semantic Shapley interaction enables our model to capture discriminative features between similar moments. Our method learns aligned and uniform representations thus eliminating the ``island" in Figure~\ref{fig:vis1}(b), 

We present two qualitative results from ActivityNet-Captions in Figure~\ref{fig:vis1}. We observe that our approach can successfully model the relationships of temporally distant moments. For instance, ``\texttt{knitting again}" in Figure~\ref{fig:vis1}(a) and ``\texttt{the second belt}" in Figure~\ref{fig:vis1}(b) are far away from their first appearances. Our method can capture the information of ``\texttt{first knitting}" and ``\texttt{the first belt}" and infer precisely the location of the target moment. Previous contrastive learning-based methods fail to capture this semantic association due to tending to perform cross-modal contrast by the strict positional principle. These models only focus on the contextual content related to the target moment. Our method demonstrates that these large volumes of unannotated moments contain rich information and modeling their relationships can improve representation learning.

\section{Conclusion and Future Work}
This paper introduces a novel semantically aligned and uniform video grounding method, Geodesic and Game Localization (G2L), which explores more semantic information by measuring the correlation between video moments through the geodesic and models the nuances of similar moments using game-theoretic interactions. By contrasting the video features and query features in the shared space, the learned bi-modal features become similar when their semantics match, and the similarity between video moments is modeled by the geodesic-guided pushing operation. Extensive experiments demonstrate that our novel training objective significantly improves the performance of existing contrastive learning-based methods in video grounding.
This paper provides a novel perspective for cross-modal contrastive learning. In the future, we intend to apply this idea to multi-modal pre-training.

\noindent \textbf{Acknowledgment.}
This paper was partially supported by NSFC (No: 62176008) and Shenzhen Science \& Technology Research Program (No: GXWD20201231165807007-20200814115301001).

{\small
\bibliographystyle{ieee_fullname}
\bibliography{egbib}
}

\end{document}


\title{$-$Supplementary Material$-$ \\ G2L: Semantically Aligned and Uniform Video Grounding \\ via Geodesic and Game Theory}

\author{
Hongxiang Li$^{1}$, Meng Cao$^{2,1}$, Xuxin Cheng$^{1}$, Yaowei Li$^{1}$, Zhihong Zhu$^{1}$, Yuexian Zou$^{1}$\footnotemark[2]\\
$^{1}$School of Electronic and Computer Engineering, Peking University \\ $^{2}$International Digital Economy Academy (IDEA)\\
{\tt\small \{lihongxiang, chengxx, zhihongzhu, ywl\}@stu.pku.edu.cn; \{mengcao, zouyx\}@pku.edu.cn}
}

\maketitle
\ificcvfinal\thispagestyle{empty}\fi


\section{Overview}

In this supplementary material, we present the following.
\begin{itemize}
    \item Axiomatic Properties of Shapley Value (Section~\ref{sec: a1}).
    \item Proof of Equation 10 (Section~\ref{sec: a2}).
\end{itemize}

\section{Axiomatic Properties of Shapley Value}\label{sec: a1}
	In this section, we mainly introduce the axiomatic properties of Shapley value. Weber \etal~\cite{weber1988probabilistic} have proved that Shapley value is the unique metric that satisfies the following axioms: \emph{Linearity}, \emph{Symmetry}, \emph{Dummy}, and \emph{Efficiency}.
	
	\textbf{Linearity Axiom.} If two independent games $u$ and $v$ can be linearly merged into one game $w(\mathcal{U}) = u(\mathcal{U}) + v(\mathcal{U})$, then the Shapley value of each player $i \in \mathcal{N}$ in the new game $w$ is the sum of Shapley values of the player $i$ in the game $u$ and $v$, which can be formulated as:
	
	\begin{equation}
		\phi_w(i|\mathcal{N}) = \phi_u(i|\mathcal{N}) + \phi_v(i|\mathcal{N})
	\end{equation}

	\textbf{Symmetry Axiom.} Considering two players $i$ and $j$ in a game $v$, if they satisfy:
	\begin{equation}
		\forall \mathcal{U} \in \mathcal{N} \setminus \{i, j\}, v(\mathcal{U} \cup \{i\}) = v(\mathcal{U} \cup \{j\})
	\end{equation}
	then $\phi_v(i|\mathcal{N}) = \phi_v(j|\mathcal{N})$.
	
	\textbf{Dummy Axiom.} The dummy player is defined as a player without interaction with other players. Formally, if a player $i$ in a game $v$ satisfies:
	\begin{equation}
		\forall \mathcal{U} \in \mathcal{N} \setminus \{i\}, v(\mathcal{U} \cup \{i\}) = v(\mathcal{U}) + v(\{i\})
	\end{equation}
	then this player is defined as the dummy player. In this way, the dummy player $i$ has no interaction with other players, \ie $v(\{i\}) = \phi_v(i|\mathcal{N})$.

	\textbf{Efficiency Axiom.} The efficiency axiom ensures that the overall reward can be assigned to all players, which can be formulated as follows:
	\begin{equation}
		\sum_{i \in \mathcal{N}} \phi_v(i) = v(\mathcal{N}) - v(\varnothing)
	\end{equation}

\section{Proof of Equation 10}\label{sec: a2}
In this section, we provide detailed proof for Equation~10 in Section~3.5.2. The semantic Shapley interaction between moment $x$ and query $y$ in video $V_i$ can be decomposed as follows:
\vspace{-0.05cm}
\begin{align}
    \mathfrak{I}([\mathcal{H}^i_{xy}]) &= \phi([\mathcal{H}^i_{xy}]|\mathcal{H}^i \setminus \mathcal{H}^i_{xy} \cup \{[\mathcal{H}^i_{xy}]\}) \nonumber\\
    &- \phi(\mathbf{h}^V_{ix}|\mathcal{H}^i \setminus \mathcal{H}^i_{xy} \cup \{\mathbf{h}^V_{ix}\}) \nonumber\\
    &- \phi(\mathbf{h}^Q_{iy}| \mathcal{H}^i \setminus \mathcal{H}^i_{xy} \cup \{\mathbf{h}^Q_{iy}\}) \\
    &= \mathop{\mathbb{E}}\limits_{C} \{\mathop{\mathbb{E}}\limits_{\mathcal{U} \subseteq \mathcal{H}^i \setminus \mathcal{H}^i_{xy}  \atop |\mathcal{U}| = C} [f(\mathcal{U} \cup \mathcal{H}^i_{xy}) - f(\mathcal{U})] \} \nonumber\\
    &- \mathop{\mathbb{E}}\limits_{C} \{\mathop{\mathbb{E}}\limits_{\mathcal{U} \subseteq \mathcal{H}^i \setminus \mathcal{H}^i_{xy}  \atop |\mathcal{U}| = C} [f(\mathcal{U} \cup \{\mathbf{h}_ix^V\}) - f(\mathcal{U})] \} \nonumber\\
    &- \mathop{\mathbb{E}}\limits_{C} \{\mathop{\mathbb{E}}\limits_{\mathcal{U} \subseteq \mathcal{H}^i \setminus \mathcal{H}^i_{xy}  \atop |\mathcal{U}| = C} [f(\mathcal{U} \cup \{\mathbf{h}^Q_{iy}\}) - f(\mathcal{U})] \}\\
    &= \mathop{\mathbb{E}}\limits_{C} \{\mathop{\mathbb{E}}\limits_{\mathcal{U} \subseteq \mathcal{H}^i \setminus \mathcal{H}^i_{xy}  \atop |\mathcal{U}| = C} [f(\mathcal{U} \cup \mathcal{H}^i_{xy}) - f(\mathcal{U} \cup \{\mathbf{h}^V_{ix}\}) \nonumber \\
    &- f(\mathcal{U} \cup \{\mathbf{h}^Q_{iy}\}) +  f(\mathcal{U})\ ]\  \}
\end{align}
 
 
{\small
\bibliographystyle{ieee_fullname}
\bibliography{egbib}
}